\documentclass[10pt,twocolumn,letterpaper]{article}

\usepackage[pagenumbers]{cvpr} 

\usepackage{float}
\usepackage[table]{xcolor}
\usepackage[ruled,vlined]{algorithm2e}
\usepackage{textcomp}
\usepackage{tikz}
\usetikzlibrary{spy}
\usepackage{subcaption} 

\definecolor{cvprblue}{rgb}{0.21,0.49,0.74}
\usepackage[pagebackref,breaklinks,colorlinks,allcolors=cvprblue]{hyperref}

\def\paperID{*****} 
\def\confName{CVPR}
\def\confYear{2026}

\title{ImprovedGS+: A High-Performance C++/CUDA Re-Implementation \\ Strategy for 3D Gaussian Splatting}

\author{Jordi Muñoz Vicente\\
Universidad de Murcia\\
Murcia, Spain\\
{\tt\small jordi.munozv@um.es }
}

\begin{document}
\maketitle
\let\thefootnote\relax\footnotetext{Project code: \url{https://github.com/jordizv/ImprovedGS-Plus}}
\begin{abstract}
Recent advancements in 3D Gaussian Splatting (3DGS) have shifted the focus toward balancing reconstruction fidelity with computational efficiency. In this work, we propose \textbf{ImprovedGS+}, a high-performance, low-level reinvention of the ImprovedGS strategy, implemented natively within the \textbf{LichtFeld-Studio} framework. By transitioning from high-level Python logic to optimized \textbf{C++/CUDA} kernels, we achieve a significant reduction in host-device synchronization and training latency. Our implementation introduces a \textbf{Long-Axis-Split (LAS)} CUDA kernel, a custom \textbf{Laplacian-based importance kernels} with Non-Maximum Suppression (NMS) for edge scores, and an adaptive \textbf{Exponential Scale Scheduler}.

Experimental results on the Mip-NeRF360 dataset demonstrate that ImprovedGS+ establishes a new Pareto-optimal front for scene reconstruction. Our 1M-budget variant outperforms the state-of-the-art MCMC baseline by achieving a \textbf{26.8\% reduction in training time} (saving ~17 minutes per session) and utilizing \textbf{13.3\% fewer Gaussians} while maintaining superior visual quality. Furthermore, our full variant demonstrates a \textbf{1.28 dB PSNR increase} over the ADC baseline with a \textbf{38.4\% reduction in parametric complexity}. These results validate ImprovedGS+ as a scalable, high-speed solution that upholds the core pillars of Speed, Quality, and Usability within the LichtFeld-Studio ecosystem.
\end{abstract}    
\section{Introduction}
\label{sec:intro}

Recent advances in 3D Gaussian Splatting (3DGS) ~\cite{kerbl3Dgaussians} have shifted the focus to balancing reconstruction quality with computational efficiency. Building upon the foundations of \textbf{ImprovedGS} strategy~\cite{deng2025improvingdensification3dgaussian} which is built upon the \textbf{TamingGS}'s work~\cite{mallick2024taming3dgshighqualityradiance}, we propose \textbf{ImprovedGS+}: a high-performance, low-level reinvention of the ImprovedGS baseline. 

While ImprovedGS provides a robust framework for densification, our implementation seeks to push the boundaries of time efficiency by transitioning from Python-based logic to a native \textbf{C$++$/CUDA} architecture. ImprovedGS+ adopts a streamlined version of its predecessor, focusing on the two most impactful components: \textbf{Long-Axis-Split (LAS)} and \textbf{Edge-Score} importance sampling. However, we go beyond a mere port of the baseline: We simplify some ImprovedGS features while we introduce subtle but critical modifications that proved to improve terminal metrics:

\begin{itemize}
    \item \textbf{Exponential Scale Scheduling}
    \item \textbf{Optimized Positional Learning}
    \item \textbf{Customized Enhanced Laplacian Masking}
\end{itemize}
 
\vspace{0.2cm}

Our implementation is developed within the \textbf{LichtFeld-Studio} \cite{lichtfeld2025} framework. By leveraging \textbf{custom GPU kernels} for both filtering and the Long-Axis-Split operation, we minimize host-device synchronization and maximize training throughput. This allows ImprovedGS+ to achieve high-efficiency scene representation without compromising the visual fidelity required for complex, large-scale environments.

\section{Methodology}
\label{sec:methodology}

\subsection{Laplacian Filter Kernel Implementation}

To generate precise structural importance maps for densification, we implemented a custom \textbf{CUDA-native pipeline}. We transition from high-level framework calls to a sequence of specialized kernels.

\vspace{0.2cm}

The pipeline follows a distilled \textbf{Canny Edge Detection} logic~\cite{canny1986computational}:
\begin{enumerate}
    \item \textbf{Preprocessing}: Grayscale conversion and a $5 \times 5$ Gaussian blur—stored in \textbf{Constant Memory} (\texttt{\_\_constant\_\_})—to filter sensor noise.
    \item \textbf{Gradient Extraction}: Parallel computation of magnitude and orientation via Sobel operators.
    \item \textbf{Non-Maximum Suppression (NMS)}: A custom thinning kernel that compares magnitudes along the gradient vector, effectively isolating geometric manifolds.
\end{enumerate}

\vspace{0.2cm}

\begin{figure}[htbp]
    \centering
    \begin{subfigure}[b]{0.35\textwidth}
        \centering
        \begin{tikzpicture}[spy using outlines={circle, magnification=3, size=4cm, connect spies, thick}]
            \node {\includegraphics[width=\textwidth]{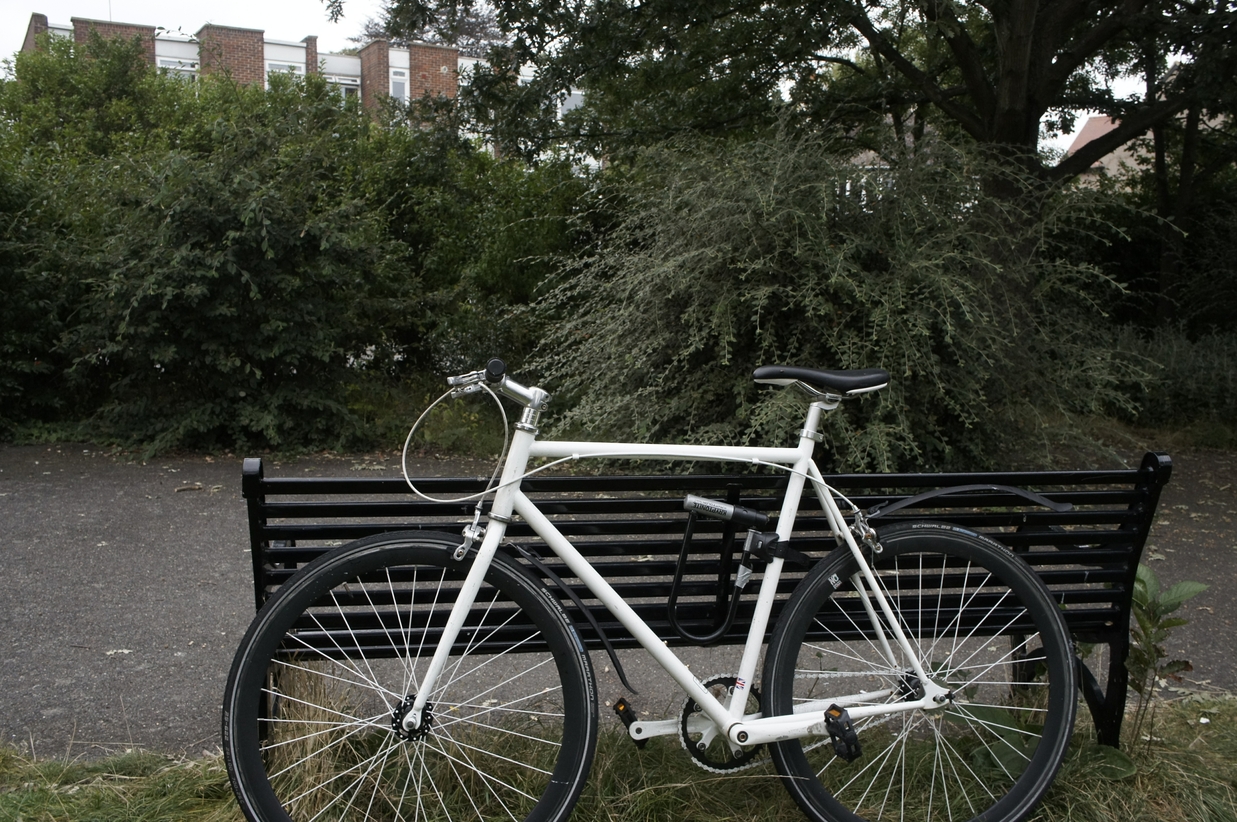}};
            \spy [red] on (-2.0,-1) in node [left] at (3.5,1.5); 
        \end{tikzpicture}
        \caption{Reference Image}
    \end{subfigure}
    \hfill
    \begin{subfigure}[b]{0.35\textwidth}
        \centering
        \begin{tikzpicture}[spy using outlines={circle, magnification=3, size=4cm, connect spies, thick}]
            \node {\includegraphics[width=\textwidth]{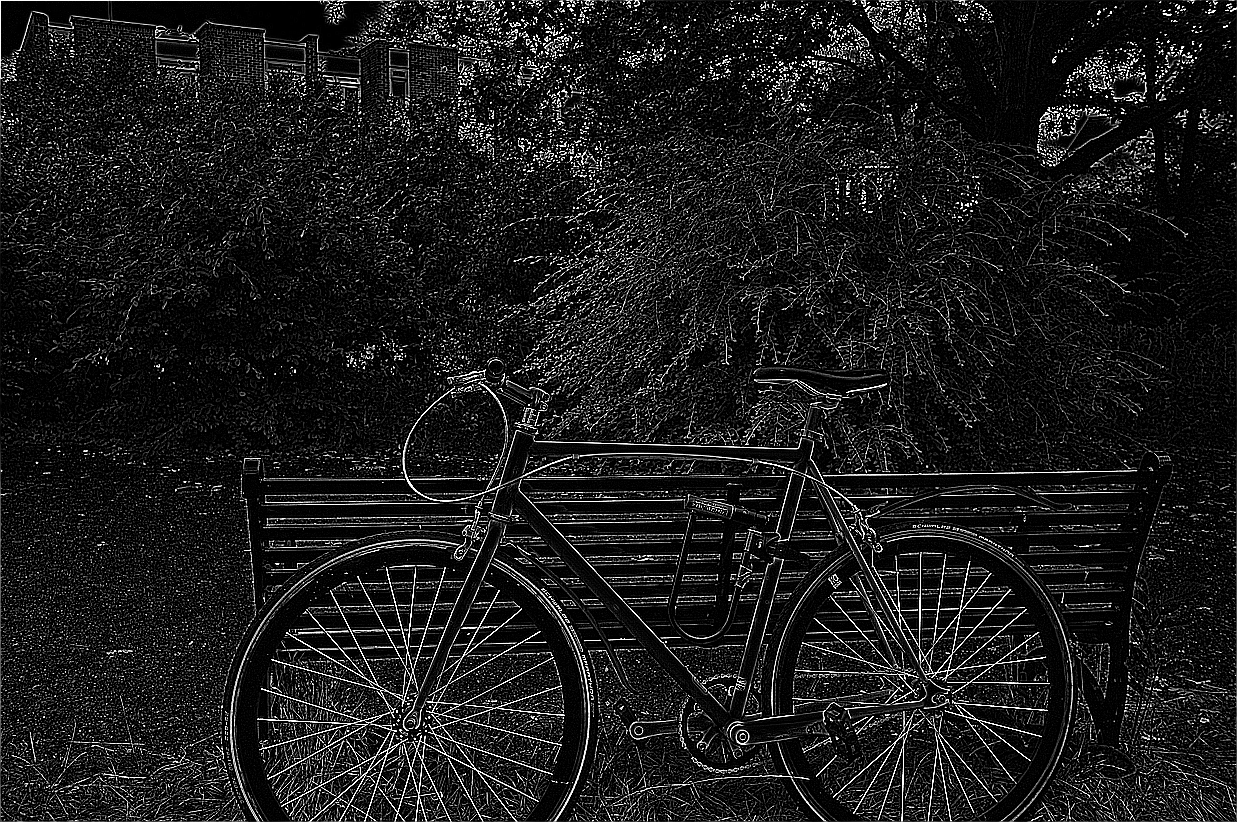}};
            \spy [red] on (-2.0,-1) in node [left] at (3.5,1.5); 
        \end{tikzpicture}
        \caption{ImprovedGS Mask (Baseline)}
    \end{subfigure}
    \hfill
    \begin{subfigure}[b]{0.35\textwidth}
        \centering
        \begin{tikzpicture}[spy using outlines={circle, magnification=3, size=4cm, connect spies, thick}]
            \node {\includegraphics[width=\textwidth]{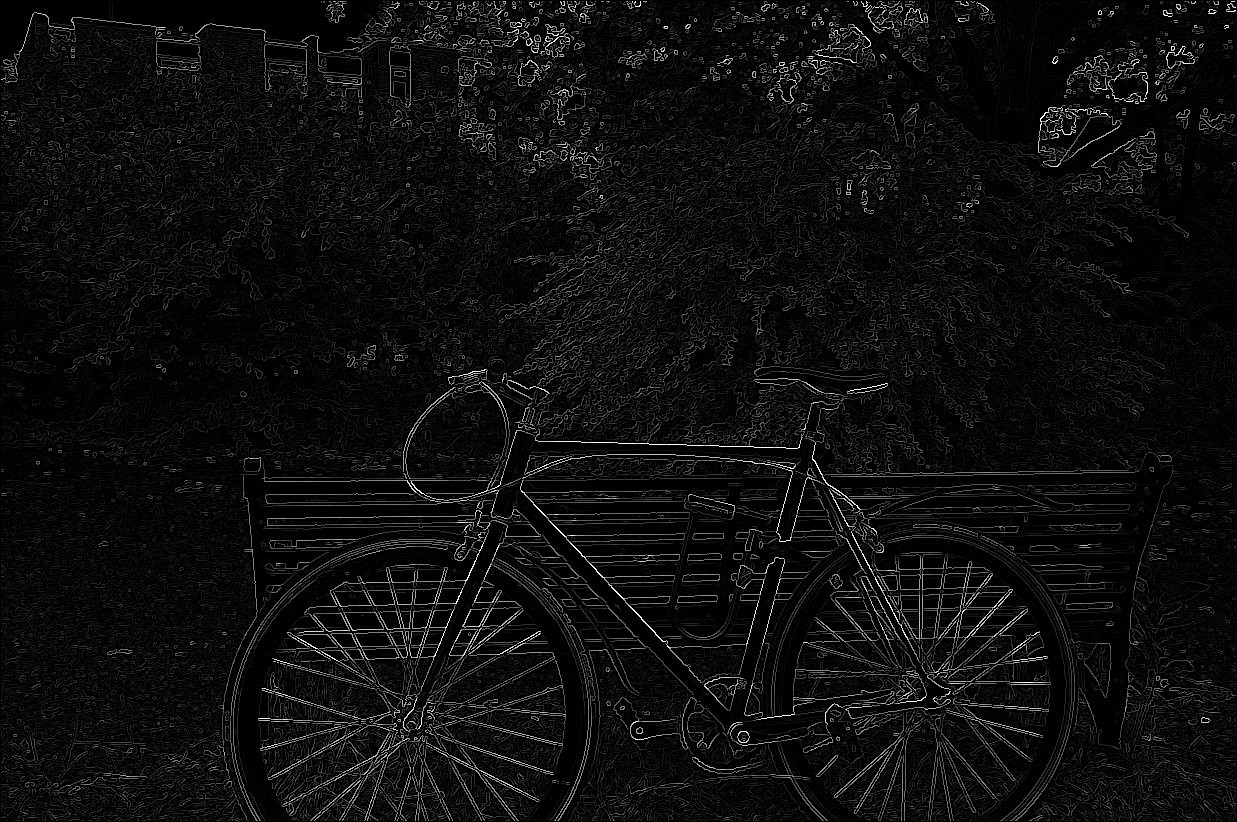}};
            \spy [red] on (-2.0,-1) in node [left] at (3.5,1.5); 
        \end{tikzpicture}
        \caption{ImprovedGS+ (Ours)}
    \end{subfigure}

    \caption{Visual comparison of edge importance maps on the \textit{Bicycle} scene. The zoomed regions highlight how our CUDA-based NMS kernel reduces the noise and focuses in a thinned structural backbone compared to the baseline.}
    \label{fig:bicycle_comparison}
\end{figure}

As illustrated in Figure~\ref{fig:bicycle_comparison}, while the Baseline (original Python-based ImprovedGS) (b) suffers from excessive stipple noise on planar surfaces, our ImprovedGS+ approach (c) produces a significantly \textbf{thinned structural backbone}. This prevents ``densification drift" by ensuring primitives are placed only on true geometric boundaries. Furthermore, we apply median-normalization to mitigate photometric outliers, ensuring high-intensity signals are preserved without being suppressed by global peaks.

\paragraph{Future Work}
Future versions could further reduce VRAM bandwidth through \textbf{Kernel Fusion}, consolidating the initial stages into a single monolithic kernel using \textbf{Shared Memory} (\texttt{\_\_shared\_\_}) for intermediate storage.

\subsection{Direct CUDA Kernel Long-Axis-Split (LAS)}

To achieve maximum efficiency during scene densification, ImprovedGS moves away from standard isotropic splitting and cloning in favor of an exclusive \textbf{Long-Axis-Split (LAS)} strategy. Within \textit{LichtFeld-Studio} framework, we implemented a custom CUDA kernel specifically for this purpose. Unlike the original Python-based implementation, which relies on high-level PyTorch function calls and multiple memory passes, our approach encapsulates the entire splitting logic within a \textbf{single CUDA kernel}. In this architecture, all geometric and attribute operations are computed in parallel at the \textbf{individual Gaussian level}.

\begin{algorithm}
\DontPrintSemicolon
\SetAlgoLined
\SetKwInOut{Input}{Data}
\Input{Gaussian attributes \textbf{$g$}, Rotation Matrix \textbf{$R$}.}
\BlankLine

\ForEach{Gaussian $g \in \text{scene}$ to split}{
    \tcp{Identify the principal axis and magnitude offset}
    $l_{idx} \gets \text{get\_max\_index}(g.\text{scale})$ \;
    $offset_{mag} \gets \exp(g.\text{scale}_{l_{idx}}) \times \alpha$ \;
    
    \BlankLine
    \tcp{Update scale components}
    $nw\_scale_{l_{idx}} \gets g.\text{scale}_{l_{idx}} + \log(\alpha)$ \;
    $nw\_scale_{other} \gets g.\text{scale}_{other} + \log(\gamma)$ \;

    \BlankLine
    \tcp{Update opacity (applying reduction factor)}
    $raw\_opac \gets \sigma(g.opac) \times \beta$ \;
    $nw\_opac \gets \sigma^{-1}(raw\_opac)$ \;

    \BlankLine
    \tcp{Compute global position offset using \textbf{$R$}}
    $\vec{v}_{offset} \gets \mathbf{R}[:, l_{idx}] \times offset_{mag}$ \;
    $nw\_pos_{0} \gets g.pos + \vec{v}_{offset}$ \;
    $nw\_pos_{1} \gets g.pos - \vec{v}_{offset}$ \;
}
\caption{Long-Axis-Split Gaussians In Place}
\end{algorithm}

It is important to note that we modify using both \textbf{storage data} (raw parameters) and \textbf{physical data} (spatial parameters retrieved via activation functions). Since scales are stored in logarithmic space, we perform the shrinking operation directly on the storage data using the logarithmic properties to ensure numerical consistency.

\paragraph{Algorithm 1} where $\alpha$ represents the shrinking factor in the longest axis (set to $0.5$ in our version) and $\gamma$ the shrinking factor for the secondary axes (set to $0.85$). Physical scale values are only retrieved via exponential activation when calculating the spatial offset. Similarly, for opacity, we reduce the physical brightness by a factor of $\beta =0.6$ before converting back to the raw logit space. 

\paragraph{Specialized Global Transformation for LAS} Finally, our custom specialized kernel CUDA optimizes the global coordinate transformation for the new Gaussian positions. Since the split displacement occurs strictly along a single local basis vector (the principal axis), the global displacement is mathematically equivalent to a linear scaling of the corresponding column of the rotation matrix.

\begin{algorithm}
\DontPrintSemicolon
\SetAlgoLined
\SetKwInOut{Input}{Input}
\SetKwInOut{Output}{Output}
\Input{Rotation matrix $\mathbf{R}$ (flat $3 \times 3$), principal axis index $l_{idx}$, physical scale $s_{phys}$}
\Output{Global displacement vector $\mathbf{d}$}
\BlankLine
$d_x \gets \mathbf{R}[l_{idx}] \times s_{phys}$ ;
\newline
$d_y \gets \mathbf{R}[l_{idx} + 3] \times s_{phys}$ ;
\newline
$d_z \gets \mathbf{R}[l_{idx} + 6] \times s_{phys}$ ;
\caption{Coordinate Transformation for LAS}
\end{algorithm}

 By extracting this column directly, we avoid the computational cost of a full $3 \times 3$ matrix-vector multiplication, reducing the operation from 9 multiplications and 6 additions to only 3 multiplications.

\subsection{Refinement Setup}

We operate under the observation that reducing densification frequency can preserve reconstruction quality while significantly minimizing computational overhead. Consequently, we keep a fixed densification interval of 500 iterations. By constraining the densification window from iteration 500 to 15,000, we restrict the model to exactly 30 total densification steps. We specifically investigate the interplay between an exclusive Long-Axis-Split (LAS) approach and the Gaussian scaling learning rate, demonstrating that our specialized learning schedule improves both initial convergence speed and terminal metrics. The strategy is divided into a two-stage process:

\begin{itemize}
    \item \textbf{Stage I: High-Momentum Expansion.} To compensate for the removal of Gaussian cloning, the \textbf{initial scale learning rate is increased  to 0.020} ($\times 4$ original initial scale rate). This allows the primitives to rapidly expand and occupy empty scene volume. When combined with the LAS primitive, this ensures that each split is spatially deterministic and non-overlapping.
    \item \textbf{Stage II: Precision Refinement.} To capture high-frequency details that a high scale learning rate might otherwise overlook, we transition into a refinement phase. Using an \textbf{Exponential Learning Rate Scheduler} with $\gamma = (0.1)$ from initial 0.020 is gradually decayed to a final value 0.002 learning rate. This stabilizes the primitives, allowing them to represent complex, fine-grained geometries, which boosts both PSNR and SSIM metrics while at the same time preventing redundant primitives, lowering the final Gaussian count.
\end{itemize}

\paragraph{Positional Learning Rate Optimization} Furthermore, we have significantly increased the positional (means) learning rate. The original ImprovedGS baseline utilizes an initial learning rate of 0.00004 decaying to 0.000002, our implementation employs an \textbf{initial rate of 0.000128} decaying to \textbf{0.0000128}. As demonstrated in our Ablation Study, although a higher positional learning rate initially faces challenges in representing the scene, it eventually unlocks higher potential in PSNR and SSIM, particularly when combined with the scale variability provided by our specialized scale scheduler.

\paragraph{Global Warm-up Phase (Initial Score Masking)}
A subtle yet functional nuance in our pipeline is the handling of early-stage importance scores. For the first three densification steps (iterations 500, 1,000, and 1,500), we suspend the gradient-threshold masking, allowing a number of Gaussians to split based purely on their calculated edge scores.

This ``warm-up" phase ensures that outdoor scenes with broad, low-gradient surfaces are sufficiently populated before the model transitions into strict structural refinement.
\definecolor{second}{HTML}{FEC088}
\definecolor{best}{HTML}{FD8488} 
\section{Experiments and Results}

This section evaluates the proposed approach  within LichtFeld-Studio framework both quantitatively and qualitatively. The evaluation was conducted using an NVIDIA RTX A4500 GPU. Results for the other techniques within LichtFeld Studio such as MCMC strategy~\cite{kheradmand20243d} and ADC (based on 3DGS), including training times, were obtained on the same hardware and environment to ensure comparability.

\subsection{Datasets and Metrics}

We evaluate our method exclusively on the \textbf{Mip-Nerf360} dataset~\cite{Barron2022}, covering a diverse range of nine scenes: \textit{Garden, Bicycle, Stump, Bonsai, Counter, Kitchen, Room, Flowers,} and \textit{Treehill}.

Although other benchmarks such \textit{as Tanks \& Temples}~\cite{Knapitsch2017} and \textit{Deep Blending}~\cite{Hedman2018} are common in the literature, we had to omit them from this study. This decision is due to the current validation architecture of the \textbf{LichtFeld-Studio} framework, which was returning metrics based on train dataset for both Tank\&Temples and DeepBlending. To ensure the most rigorous and consistent comparative analysis within this ecosystem, we focused on the comprehensive Mip-NeRF360 suite, which provides a balanced mix of complex outdoor environments and structured indoor manifolds.

We compare the common quality metrics: peak signal-to-noise ratio (PSNR) and structural similarity (SSIM). It is important to note that our work focuses in resource efficiency: Aiming to achieve high quality with low resource usage. We assess these qualities by timing the optimization (Train time) in minutes and seconds and by counting the final number of Gaussians (\#G) in millions ($\times10^{6}$) for clarity.

\subsection{Ablation Study}
To validate the individual components of the \textbf{ImprovedGS+} (IGS+) framework, we conduct a comparative analysis across two distinct environments from the Mip-Nerf360 dataset: \textit{Garden} (Outdoor) and \textit{Bonsai} (Indoor). 

We evaluate three configurations without any budget to isolate the impact of our optimizations:
\begin{itemize}
    \item \textbf{IGS+ (Ours)}: The full implementation featuring retuned \texttt{position\_lr} boundaries and init scale rate along with our Exponential Scale Scheduling.
    \item \textbf{w/o Scale Scheduler}: The full IGS+ pipeline but without the exponential decay for Gaussian scales.
    \item \textbf{IGS (Baseline)}: Our C++/CUDA re-implementation of the original ImprovedGS, using its native positional and scale learning rates.
\end{itemize}

\begin{table}
    \centering
    \scalebox{0.62}{
        \begin{tabular}{l| ccc | ccc }
        \hline
        \textbf{Scene} & \multicolumn{3}{c|}{Garden} & \multicolumn{3}{c|}{Bonsai} \\
        \hline
        \textbf{} & PSNR $\uparrow$ & SSIM$\uparrow$ & $\#G(10^{6})$ $\downarrow$ & PSNR $\uparrow$ & SSIM$\uparrow$ & $\#G(10^{6})$ $\downarrow$ \\
        \hline
         \textbf{IGS (Baseline)} & 26.86 & 0.845 & 1.310 & 30.39 & 0.935 & 0.535 \\ 
        \hline
        \textbf{IGS+ (Ours)} & 26.08 & 0.809 & 1.400 & 30.34 & 0.932 & 0.573 \\
        \hline
        \textbf{-w/o scale scheduler} & 25.95 & 0.806 & 1.400 & 30.18 & 0.933 & 0.575 \\
        \hline
    \end{tabular}
    }
    \caption{Bonsai and Garden scenes Ablations table at 7,000 iterations for ImprovedGS+. Averaged from 6 independent runs.}
    \label{tab:ablations_7k}
\end{table}

At 7,000 iterations (Table~\ref{tab:ablations_7k}), the standard IGS baseline maintains a slight lead. This is an intentional design choice: our higher \textit{positional\_lr} at Stage I expansion prioritizes the rapid acquisition of geometric volume and primitive displacement over immediate photometric precision. This trade-off is critical for preventing the ``under-reconstruction" of complex areas.

By the terminal state at 30,000 iterations (Table~\ref{tab:ablations_30k}), the benefits of our LichtFeld-Studio parameter tuning become clear:
\begin{enumerate}
    \item \textbf{Garden}: PSNR improves from 29.05 (Baseline) to \textbf{29.39} (Ours).
    \item \textbf{Bonsai}: PSNR improves from 33.16 (Baseline) to \textbf{33.37} (Ours).
\end{enumerate}

As shown in Table~\ref{tab:ablations_30k}, in \textit{Garden} the IGS Baseline tops at 2.61M Gaussians, unable to capture further complexity. Our \texttt{positional\_lr} allows for a more adaptive densification, reaching 2.93M Gaussians, which is directly responsible for the 0.34 dB PSNR improvement.

\paragraph{Impact of the Scale Scheduler}
The removal of the Exponential Scale Scheduler (\textit{w/o scale scheduler}) leads to a measurable drop in terminal metrics. In the \textit{Bonsai} scene, disabling the scheduler results in a \textbf{0.42 dB} drop in PSNR. This confirms that our scheduler is essential for ``freezing" the structural details and preventing the primitives from oscillating or over-expanding during the final refinement phase.

\begin{table}
    \centering
    \scalebox{0.62}{
        \begin{tabular}{l| ccc | ccc }
        \hline
        \textbf{Scene} & \multicolumn{3}{c|}{Garden} & \multicolumn{3}{c|}{Bonsai} \\
        \hline
        \textbf{} & PSNR $\uparrow$ & SSIM$\uparrow$ & $\#G(10^{6})$ $\downarrow$ & PSNR $\uparrow$ & SSIM$\uparrow$ & $\#G(10^{6})$ $\downarrow$ \\
        \hline
        \textbf{IGS (Baseline)} & 29.05 & 0.887 & 2.614 & 33.16 & 0.953 & 0.827 \\ 
        \hline
        \textbf{IGS+ (Ours)} & 29.39 & 0.896 & 2.930 & 33.37 & 0.954 & 0.830 \\
        \hline
        \textbf{-w/o scale scheduler} & 29.16 & 0.894  & 2.937 & 32.95 & 0.952 & 0.843 \\
        \hline
    \end{tabular}
    }
    \caption{Bonsai and Garden scenes Ablations table at 30,000 iterations for ImprovedGS+. Averaged from 6 independent runs.}
    \label{tab:ablations_30k}
\end{table}

\subsection{Results}

\begin{table}[H]
    \scalebox{0.82}{
    \begin{tabular}{l| ccccc}
        \hline
        \textbf{Dataset} & \multicolumn{4}{c}{\textbf{MipNerf360}} \\
        \hline
         & \textbf{PSNR} $\uparrow$ & \textbf{SSIM} $\uparrow$ & \textbf{Time} $\downarrow$ & \textbf{$\#G(10^{6})$} $\downarrow$ \\
        \hline
        \textbf{MCMC 1M} & 28.64 & 0.848 & 62m14s & 1.000 \\
        \hline
        \textbf{Ours (IGS+ 1M) } & \cellcolor{best!70}28.79 & \cellcolor{best!70}0.859 & \cellcolor{best!70} 45m33s & \cellcolor{best!70}0.867 \\
        \hline
        \textbf{Ours (IGS+) } & 29.38 & 0.877 & 62m10s & 1.688 \\
        \hline
        \textbf{ADC}  & 28.10 & 0.848 & 76m02s & 2.741 \\
        \hline

        \hline
    \end{tabular}
    }
    \caption{Experimental Results across the Mip-Nerf360 dataset at 30,000 iterations. Metrics for each scene are averaged from 6 independent runs. For a fair comparative analysis, the best results between MCMC 1M and ImprovedGS+ 1M are highlighted in red.}
    \label{tab:results}
\end{table}
\section{Conclusion}

As summarized in Table~\ref{tab:results}, \textbf{ImprovedGS+ (IGS+)} demonstrates clear dominance over existing strategies within the \textbf{LichtFeld-Studio} framework. Compared to the State-Of-The-Art MCMC strategy, under an identical 1 million Gaussian budget our method achieves superior metrics in significantly less time. We report an aggregate reduction of approximately 17 minutes across all nine benchmark scenes (\textbf{26.8\% time reduction}) while utilizing \textbf{13.3\% fewer Gaussians}. This underscores the high level of optimization and ``per-primitive contribution" that IGS+ introduces to the densification process.

A key strength of our approach is its \textbf{adaptative efficiency}. As evidenced in the indoor scenes (see Appendix), IGS+ consistently reaches SOTA metrics without exhausting the maximum permitted budget. In the \textit{Room} scene, for instance, our model achieves superior PSNR using only 0.591M Gaussians compared to the 1.000M required by MCMC. Even when run without a strict budget (utilizing a 3M pre-allocated capacity), IGS+ naturally optimizes the scene representation to an average of 1.688M Gaussians — a \textbf{40\% reduction compared to the ADC strategy}. Crucially, this high-fidelity variant completes training in the same temporal window as the MCMC 1M baseline, suggesting significantly higher architectural throughput.

While the baseline restrictions observed in our ablation study initially limited growth in complex  environments, our refined \texttt{positional\_lr} scheduling successfully pushed these boundaries. By allowing for a more granular movement and precise placement of primitives (especially at the end phase of training), we directly translated increased Gaussian counts into improved visual fidelity in challenging outdoor scenes like \textit{Garden}

Ultimately, \textbf{ImprovedGS+} provides a highly scalable solution: the full version maximizes visual quality for high-end rendering, while the 1M budget variant offers a high-speed, mobile-ready alternative that performs the MCMC baseline across all metrics. This performance reinforces the core pillars of \textbf{Speed}, \textbf{Quality} and \textbf{Usability}, that define the LichtFeld-Studio ecosystem.
{
    \small
    \bibliographystyle{ieeenat_fullname}
    \bibliography{Bibliografia/references}
}

\definecolor{second}{HTML}{FEC088}
\definecolor{best}{HTML}{FD8488} 
\appendix
\clearpage
\section{Appendix}

\subsection{Quantitative Comparison per Scene}

Comparison highlight is done only between MCMC and ImprovedGS+ strategies when both using same budget.

\subsubsection*{MipNeRF360: Indoor Scenes}

\begin{table}[H]
    \centering
    \small
    \begin{tabular}{l cccc | cccc}
        \toprule
        \textbf{Scene} & \multicolumn{4}{c|}{\textbf{Bonsai}} & \multicolumn{4}{c}{\textbf{Counter}} \\
        \cmidrule(lr){2-5} \cmidrule(lr){6-9}
        & PSNR $\uparrow$ & SSIM $\uparrow$ & Time $\downarrow$ & \#G $\downarrow$ & PSNR $\uparrow$ & SSIM $\uparrow$ & Time $\downarrow$ & \#G $\downarrow$ \\
        \midrule
        \textbf{MCMC 1M} & 32.91 & 0.951 & 5m54s & 1.000 & 30.84 & 0.928 & 6m48s & 1.000 \\
        \textbf{ImprovedGS+} & \cellcolor{best!70}33.37 & \cellcolor{best!70}0.954 & \cellcolor{best!70}4m35s & \cellcolor{best!70}0.830 & \cellcolor{best!70}31.05 & \cellcolor{best!70}0.929 & \cellcolor{best!70}4m13s & \cellcolor{best!70}0.618 \\
        \textbf{ADC} & 32.10 & 0.942 & 4m12s & 1.147 & 29.19 & 0.897 & 4m08s & 0.875 \\
        \midrule \midrule 
        \textbf{Scene} & \multicolumn{4}{c}{\textbf{Kitchen}} & \multicolumn{4}{c}{\textbf{Room}} \\
        \cmidrule(lr){2-5} \cmidrule(lr){6-9}
        & PSNR $\uparrow$ & SSIM $\uparrow$ & Time $\downarrow$ & \#G $\downarrow$ & PSNR $\uparrow$ & SSIM $\uparrow$ & Time $\downarrow$ & \#G $\downarrow$ \\
        \midrule
        \textbf{MCMC 1M} & 32.43 & 0.938 & 6m57s & 1.000 & 34.24 & 0.946 & 5m00s & 1.000 \\
        \textbf{ImprovedGS+} & \cellcolor{best!70}32.47 & 0.938 & \cellcolor{best!70}5m09s & \cellcolor{best!70}0.760 & \cellcolor{best!70}34.36 & 0.946 & \cellcolor{best!70}3m16s & \cellcolor{best!70}0.591 \\
        \textbf{ADC} & 31.16 & 0.927 & 5m36s & 1.194 & 32.97 & 0.933 & 4m05s & 1.127 \\
        \bottomrule
    \end{tabular}
    \caption{Extended evaluation metrics for all Indoor scenes. Each scene averaged from 6 independent runs.}
\end{table}

\subsubsection*{MipNeRF360: Outdoor Scenes}

\begin{table}[H]
    \centering
    \small
    \begin{tabular}{l cccc | cccc}
        \toprule
        \textbf{Scene} & \multicolumn{4}{c|}{\textbf{Garden}} & \multicolumn{4}{c|}{\textbf{Bicycle}} \\
        \cmidrule(lr){2-5} \cmidrule(lr){6-9}
        & PSNR $\uparrow$ & SSIM $\uparrow$ & Time $\downarrow$ & \#G $\downarrow$ & PSNR $\uparrow$ & SSIM $\uparrow$ & Time $\downarrow$ & \#G $\downarrow$ \\
        \midrule
        \textbf{MCMC 1M} & 28.27 & 0.866 & 7m53s & 1.000 & 25.07 & 0.790 & 6m59s & 1.000 \\
        \textbf{ImprovedGS+ 1M} & \cellcolor{best!70}28.47 & \cellcolor{best!70}0.871 & \cellcolor{best!70}6m26s & 1.000 & \cellcolor{best!70}25.39 & \cellcolor{best!70}0.798 & \cellcolor{best!70}5m33s & 1.000 \\
        \textbf{ImprovedGS+} & 29.39 & 0.896 & 11m39s & 2.930 & 26.88 & 0.813 & 9m45s & 3.000 \\
        \textbf{ADC} & 28.46 & 0.870 & 13m42s & 4.082 & 25.68 & 0.803 & 14m06s & 5.211 \\
        \midrule \midrule 
        \textbf{Scene} & \multicolumn{4}{c}{\textbf{Flowers}} & \multicolumn{4}{c}{\textbf{Treehill}} \\
        \cmidrule(lr){2-5} \cmidrule(lr){6-9}
        & PSNR $\uparrow$ & SSIM $\uparrow$ & Time $\downarrow$ & \#G $\downarrow$ & PSNR $\uparrow$ & SSIM $\uparrow$ & Time $\downarrow$ & \#G $\downarrow$ \\
        \midrule
        \textbf{MCMC 1M} & \cellcolor{best!70}23.43 & \cellcolor{best!70}0.741 & 7m44s & 1.000 & 22.91 & 0.727 & 7m59s & 1.000 \\
        \textbf{ImprovedGS+ 1M} & 23.39 & 0.720 & \cellcolor{best!70}5m42s & 1.000 & \cellcolor{best!70}23.21 & \cellcolor{best!70}0.738 & \cellcolor{best!70}5m22s & 1.000 \\
        \textbf{ImprovedGS+} & 24.35 & 0.761 & 8m32s & 2.250 & 24.24 & 0.790 & 7m41s & 2.250 \\
        \textbf{ADC} & 23.32 & 0.718 & 10m42s & 3.542 & 21.86 & 0.693 & 8m32s & 3.213 \\
        \midrule \midrule 
        \textbf{Scene} & \multicolumn{4}{c}{\textbf{Stump}} \\
        \cmidrule(lr){2-5}
        & PSNR $\uparrow$ & SSIM $\uparrow$ & Time $\downarrow$ & \#G $\downarrow$ \\
        \textbf{MCMC 1M} & \cellcolor{best!70}27.65 & 0.832 & 7m00s & 1.000 \\
        \textbf{ImprovedGS+ 1M} & 27.38 &  \cellcolor{best!70}0.835 & \cellcolor{best!70}5m17s & 1.000 \\
        \textbf{ImprovedGS+} & 28.36 & 0.864 & 7m20s & 1.960 \\
        \textbf{ADC} & 28.18 & 0.848 & 10m52s & 4.280 \\
        \bottomrule
    \end{tabular}
    \caption{Extended evaluation metrics for all Outdoor scenes. Each scene averaged from 6 independent runs.}
\end{table}

\end{document}